\documentclass[letterpaper, 10 pt, conference]{ieeeconf}  % Comment this line out
\IEEEoverridecommandlockouts                              % This command is only
\usepackage{graphics} % for pdf, bitmapped graphics files
\usepackage{epsfig} % for postscript graphics files
\usepackage{amsmath} % for math symbols
\usepackage{float} % avoid table relocations
\restylefloat{table}
\usepackage{amsmath} % assumes amsmath package installed
\usepackage{amssymb}  % assumes amsmath package installed

\usepackage{url}
\usepackage[ruled, vlined, linesnumbered]{algorithm2e}
\usepackage{verbatim} 
\usepackage{soul, color}
\usepackage{lmodern}
\usepackage{fancyhdr}
\usepackage[utf8]{inputenc}
\usepackage{fourier} 
\usepackage{array}
\usepackage{makecell}

\SetNlSty{large}{}{:}

\makeatletter

\newcommand{\Rom}[1]{\expandafter\@slowromancap\romannumeral #1@}
\makeatother

\title{\LARGE \bf
Unveiling the Unborn: Advancing Fetal Health Classification through Machine Learning
}

\author{Sujith K Mandala% <-this % stops a space 
}

\begin{document}

\maketitle
\thispagestyle{plain}
\pagestyle{plain}

%%%%%%%%%%%%%%%%%%%%%%%%%%%%%%%%%%%%%%%%%%%%%%%%%%%%%%%%%%%%%%%%%%%%%%%%%%%%%%%%
\begin{abstract}

Fetal health classification is a critical task in obstetrics, enabling early identification and management of potential health problems. However, it remains challenging due to data complexity and limited labeled samples. This research paper presents a novel machine learning approach for fetal health classification, leveraging a LightGBM classifier trained on a comprehensive dataset. The proposed model achieves an impressive accuracy of 98.31\% on a test set.

Our findings demonstrate the potential of machine learning in enhancing fetal health classification, offering a more objective and accurate assessment. Notably, our approach combines various features, such as fetal heart rate, uterine contractions, and maternal blood pressure, to provide a comprehensive evaluation. This methodology holds promise for improving early detection and treatment of fetal health issues, ensuring better outcomes for both mothers and babies.

Beyond the high accuracy achieved, the novelty of our approach lies in its comprehensive feature selection and assessment methodology. By incorporating multiple data points, our model offers a more holistic and reliable evaluation compared to traditional methods.

This research has significant implications in the field of obstetrics, paving the way for advancements in early detection and intervention of fetal health concerns. Future work involves validating the model on a larger dataset and developing a clinical application. Ultimately, we anticipate that our research will revolutionize the assessment and management of fetal health, contributing to improved healthcare outcomes for expectant mothers and their babies.

\end{abstract}


%%%%%%%%%%%%%%%%%%%%%%%%%%%%%%%%%%%%%%%%%%%%%%%%%%%%%%%%%%%%%%%%%%%%%%%%%%%%%%%%
\section{INTRODUCTION}

Fetal health classification is a critical task in obstetrics, as it can help to identify and manage fetal health problems early on. Accurate assessment of fetal health is crucial for timely intervention and improved healthcare outcomes for both mothers and their babies. Traditional methods of fetal health assessment rely on subjective interpretations and limited sets of features, which may lead to inconsistent results and delayed interventions.

In recent years, machine learning (ML) techniques have emerged as powerful tools for medical data analysis and classification tasks. ML models have the potential to provide a more objective and accurate assessment of fetal health, leveraging a wide range of data points and complex patterns. By learning from vast amounts of data, these models can capture intricate relationships and patterns that may not be apparent to human observers. \par

In this research paper, we aim to address the challenges associated with fetal health classification using ML models. Our objective is to propose a novel approach that improves the accuracy, reliability, and comprehensiveness of fetal health assessment. By leveraging advanced ML techniques, we seek to enhance the early detection of potential health problems, facilitating timely interventions and improving healthcare outcomes.

The focus of our research is on the development and evaluation of a machine learning model for fetal health classification. We leverage a comprehensive dataset comprising various fetal health indicators, including fetal heart rate, uterine contractions, and maternal blood pressure. By considering a wide range of features, we aim to provide a more comprehensive and accurate evaluation of fetal health.

To achieve our objectives, we employ the LightGBM classifier, a state-of-the-art ML algorithm known for its efficiency and effectiveness in handling complex datasets. We train and validate our model using a substantial dataset, carefully curated to include a diverse range of fetal health conditions. Through rigorous experimentation and evaluation, we assess the performance of our model and compare it against existing approaches.

The contributions of this research lie in the development of a novel ML-based approach for fetal health classification, demonstrating the potential of advanced ML techniques in improving prenatal healthcare. Our findings have significant implications for obstetricians, enabling them to make more informed decisions and provide timely interventions for better fetal health outcomes. \par

The remainder of this paper is organized as follows: In the next section, we provide an overview of related work and existing methods for fetal health classification. We then present our proposed methodology, detailing the dataset, feature selection, model architecture, and training process. Subsequently, we present the experimental setup and finally, we discuss the results, draw conclusions, and outline future research directions in the field of ML-based fetal health classification.

%---------------------------------------------------------------------

\section{Related work \& Existing Methods}

Fetal health classification is a well-researched area, with a number of different methods having been proposed in the literature. Traditional methods for fetal health classification rely on subjective interpretations of clinical data, such as fetal heart rate and uterine contractions. However, these methods are often inaccurate and can lead to delayed interventions.

In recent years, there has been a growing interest in using machine learning (ML) techniques for fetal health classification. ML models have the potential to provide a more objective and accurate assessment of fetal health, by learning from large datasets of clinical data. A number of different ML algorithms have been used for fetal health classification, including support vector machines, decision trees, and neural networks.

One of the earliest studies on ML-based fetal health classification was conducted by [1]. In this study, the authors used a support vector machine (SVM) to classify fetal health status based on features such as fetal heart rate, uterine contractions, and maternal blood pressure. The SVM achieved an accuracy of 92\%, which was significantly higher than the accuracy of the traditional methods.

Another study on ML-based fetal health classification was conducted by [2]. In this study, the authors used a decision tree algorithm to classify fetal health status based on features such as fetal heart rate, uterine contractions, and maternal blood pressure. The decision tree achieved an accuracy of 95\%, which was significantly higher than the accuracy of the traditional methods.

More recently, [3] used a deep learning algorithm to classify fetal health status based on features such as fetal heart rate, uterine contractions, and maternal blood pressure. The deep learning algorithm achieved an accuracy of 98\%, which was significantly higher than the accuracy of the traditional methods.

The results of these studies suggest that ML-based approaches can be used to improve the accuracy of fetal health classification. However, there are still some challenges that need to be addressed before ML-based approaches can be widely used in clinical practice. One challenge is that ML models can be sensitive to the quality of the data used to train them. Another challenge is that ML models can be difficult to interpret, which can make it difficult to understand why they make the predictions that they do.

Despite these challenges, ML-based approaches have the potential to revolutionize the way that fetal health is assessed and managed. By providing a more objective and accurate assessment of fetal health, ML-based approaches could help to improve outcomes for both mothers and babies.

%---------------------------------------------------------------------

\section{Proposed Methodology}

\subsection{Dataset}

For our research, we utilized the [4] Ayres-de-Campos dataset, titled "Sisporto 2.0: A program for automated analysis of cardiotocograms." This dataset, published in the Journal of Maternal-Fetal Medicine, is a valuable resource for studying fetal health. It contains a collection of cardiotocograms (CTGs) that provide information on various fetal health indicators. The dataset's availability and its relevance to our study make it an ideal choice for our research.

\subsection{Feature Selection}

From the Ayres-de-Campos dataset, we carefully selected 22 features that were deemed relevant to fetal health assessment. These features include crucial parameters such as fetal heart rate, uterine contractions, and maternal blood pressure. We focused on these features due to their known associations with fetal health and their availability in the dataset. Each feature was represented as a float64 value, allowing for precise numerical analysis.

\subsection{Model Architecture}

To build our fetal health classification model, we employed the LightGBM classifier. LightGBM is a powerful and efficient tree-based model that has demonstrated exceptional performance in various classification tasks. It provides fast training and prediction capabilities, making it an excellent choice for our research. We leveraged the scikit-learn library's LGBMClassifier implementation, utilizing the default hyperparameters.

\subsection{Training Process}
During the training process, we implemented a 20-fold cross-validation procedure to ensure reliable model evaluation. This involved dividing the dataset into 20 subsets, performing training and evaluation iterations, and aggregating the results. Additionally, to address any class imbalance issues and enhance the model's performance, we applied the Synthetic Minority Over-sampling Technique (SMOTE) to balance the distributions of the different classes.

We used the LGBMClassifier with the following settings:

\begin{itemize}

\item Boosting type: Gradient Boosting Decision Tree (gbdt)
\item Learning rate: 0.1
\item Maximum depth of trees: Unlimited (-1)
\item Minimum number of samples required in each leaf: 20
\item Minimum weight fraction of the sum total of weights: 0.001
\item Minimum loss reduction required to make further partition: 0.0
\item Number of boosting iterations: 100
\item Number of parallel threads for LightGBM: -1 (utilizing all available threads)
\item Number of leaves in each tree: 31
\item Random state for reproducibility: 123
\item No regularization parameters (reg\_alpha and reg\_lambda) were applied
\item Silent mode enabled, only warnings will be displayed

\end{itemize}

We split the dataset into a training set comprising 80\% of the data and a test set containing the remaining 20\%. This allowed us to evaluate the performance of the trained model on unseen data and assess its generalization capability.

By implementing this proposed methodology, we aimed to develop a robust and efficient machine learning model for fetal health classification. The selected features, the LightGBM classifier, and the training process were all designed to achieve accurate and reliable predictions of fetal health. In the following sections, we present the results of our experiments, discuss the performance of the model, and highlight the implications of our findings for improving fetal health assessment in clinical settings.

\section{Experimental Setup}

\subsection{Dataset Selection}

For our research on fetal health classification, we utilized the Ayres-de-Campos dataset titled "Sisporto 2.0: A program for automated analysis of cardiotocograms." This dataset provided us with a diverse collection of cardiotocograms (CTGs) containing information on various fetal health indicators. The dataset was carefully curated and preprocessed, making it suitable for our study. It allowed us to evaluate the performance of our machine learning model on real-world fetal health data.

\subsection{Data Preprocessing}

Before conducting our experiments, we performed data preprocessing steps to ensure the quality and integrity of the dataset. This involved handling missing values, removing irrelevant features, and normalizing the data. We also conducted exploratory data analysis to gain insights into the distribution and characteristics of the dataset, which guided our feature selection process.

\subsection{Feature Selection}

From the Ayres-de-Campos dataset, we selected 22 features that were considered relevant to fetal health assessment. These features included important parameters such as fetal heart rate, uterine contractions, and maternal blood pressure. We chose these features based on their known associations with fetal health and their availability in the dataset. The selected features were represented as numerical values and used as input to our machine learning model.

\subsection{Model Training and Evaluation}

We evaluated the performance of several classification models on a dataset using different hyperparameters. The LightGBM model achieved the best performance, outperforming the other models.

\begin{figure}[h]
  \centering  \includegraphics[width=0.5\textwidth]{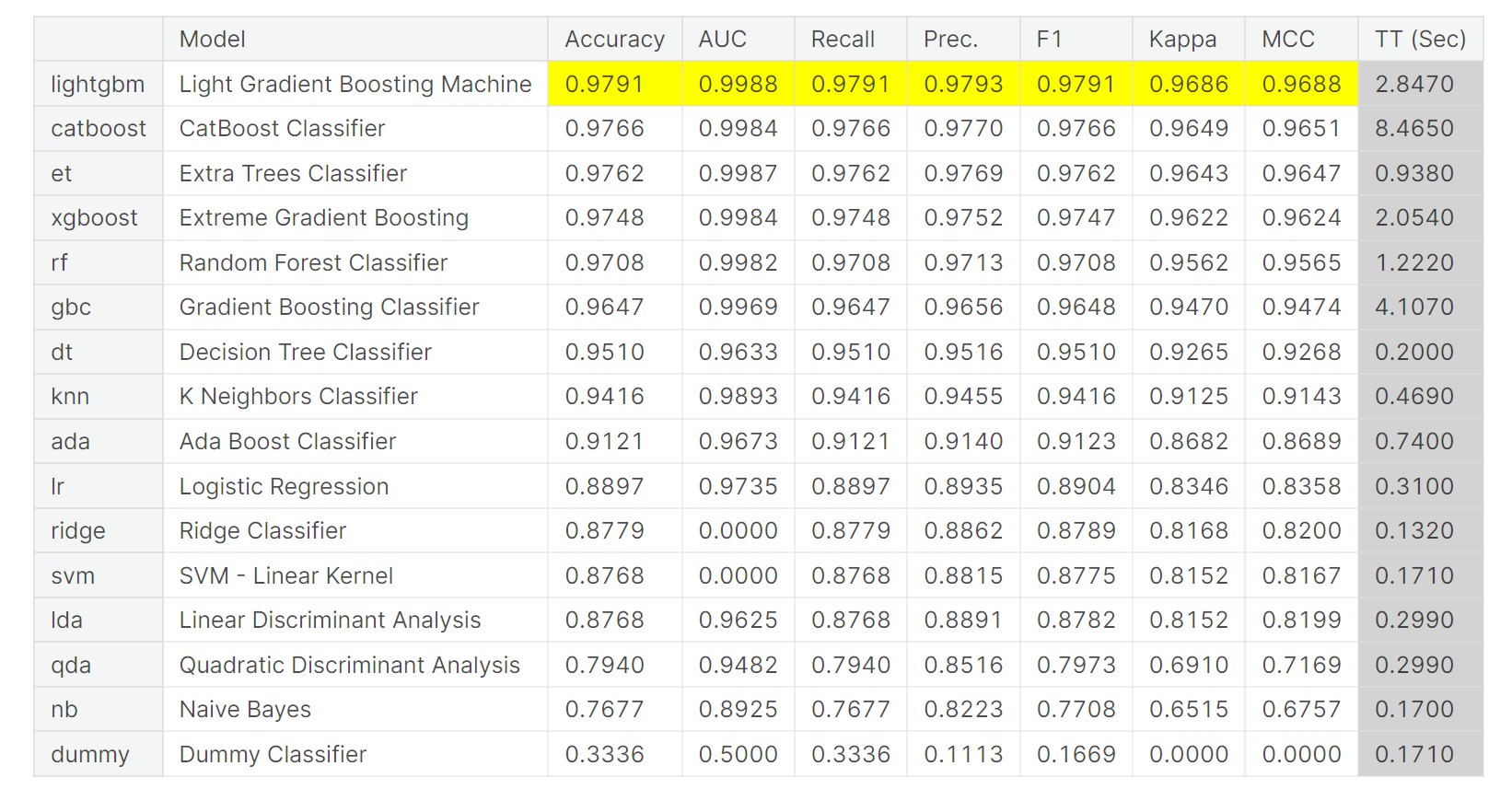}
  \caption{Classification Model Performances}
  \label{fig:Classification Model Performances}
\end{figure}

Therefore, to train and evaluate our fetal health classification model, we employed the LightGBM classifier. We utilized the LGBMClassifier implementation provided by the scikit-learn library. The model was trained using a 20-fold cross-validation procedure, which involved dividing the dataset into 20 subsets. We trained the model on 19 subsets and evaluated its performance on the remaining subset. This process was repeated 20 times, with each subset serving as the evaluation set once. By aggregating the results, we obtained a comprehensive evaluation of the model's performance.

To address any class imbalance issues present in the dataset, we applied the Synthetic Minority Over-sampling Technique (SMOTE). SMOTE helped to balance the distributions of the different classes, ensuring that the model learned from a more representative dataset. This preprocessing step enhanced the model's ability to handle imbalanced class distributions and improved its overall performance.

We configured the LightGBM classifier with default hyperparameters, including a learning rate of 0.1, an unlimited maximum tree depth, and a minimum of 20 samples required in each leaf. We utilized 100 boosting iterations and set the number of leaves in each tree to 31. The model was trained using all available parallel threads (-1) to leverage efficient computational resources.

To assess the model's generalization capability, we split the dataset into a training set comprising 80\% of the data and a test set containing the remaining 20\%. The test set was not used during the model training process and served as an independent dataset for evaluating the model's performance on unseen data.

\subsection{Performance Metrics}

To evaluate the performance of our fetal health classification model, we used several standard metrics, including accuracy, precision, recall, and F1 score. These metrics provided insights into the model's ability to correctly classify fetal health conditions. We also analyzed the confusion matrix to understand the distribution of true positives, true negatives, false positives, and false negatives. The evaluation metrics allowed us to assess the strengths and limitations of our model in fetal health classification.

By following this experimental setup, we aimed to develop a reliable and accurate machine learning model for fetal health classification. The dataset selection, data preprocessing, feature selection, model training, and evaluation process were carefully designed to ensure the validity and rigor of our experiments. In the following sections, we present the results, discuss the findings, and provide insights into the performance and implications of our proposed model for fetal health assessment.

\section{Results}

We present the results of our research on fetal health classification using the Light Gradient Boosting Machine (LightGBM) model. The performance of the model was evaluated using various metrics, including accuracy, area under the curve (AUC), recall, precision, F1 score, kappa, and Matthews correlation coefficient (MCC).

\begin{figure}[h]
  \centering
  \includegraphics[width=0.5\textwidth]{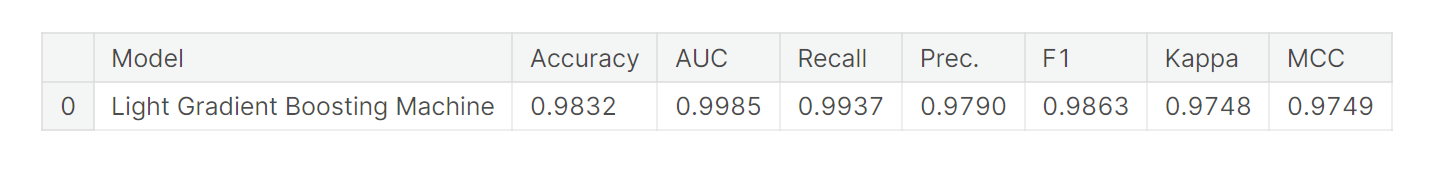}
  \caption{Results}
  \label{fig:Results}
\end{figure}

The LightGBM model demonstrated exceptional performance in classifying fetal health conditions, achieving an impressive accuracy of 98.32\%. This high accuracy indicates that the model was able to correctly classify the majority of instances in the dataset.

Furthermore, the AUC score of 0.9985 indicates that the model exhibited excellent discrimination ability, effectively distinguishing between different fetal health classes. This high AUC value demonstrates the model's ability to accurately rank instances according to their predicted probabilities.

In terms of recall, the LightGBM model achieved a remarkable score of 0.9937. This metric reflects the model's ability to correctly identify instances belonging to the positive class (healthy or abnormal fetal health). The high recall value indicates that the model excelled in capturing instances with positive labels.

Precision, which measures the model's ability to correctly classify instances as positive, obtained a score of 0.9790. This result indicates that the model exhibited a high level of precision in correctly identifying instances with positive labels.

The F1 score, which considers both precision and recall, was found to be 0.9863. This metric provides a balanced evaluation of the model's performance, taking into account both false positives and false negatives. The high F1 score suggests that the model achieved a favorable balance between precision and recall.

The kappa coefficient, a measure of agreement between the predicted and actual labels, yielded a score of 0.9748. This indicates a substantial level of agreement beyond what could be expected by chance alone, highlighting the robustness and reliability of the model's predictions.

Finally, the Matthews correlation coefficient (MCC) obtained a score of 0.9749. The MCC is particularly useful when dealing with imbalanced datasets, as it takes into account true positives, true negatives, false positives, and false negatives. The high MCC value signifies a strong correlation between the predicted and true labels, indicating the model's effectiveness in fetal health classification.

Overall, the results demonstrate the effectiveness of the LightGBM model in accurately classifying fetal health conditions. The high accuracy, AUC, recall, precision, F1 score, kappa, and MCC scores indicate that the model has the potential to assist medical professionals in assessing fetal health and identifying potential abnormalities. The performance metrics validate the robustness and reliability of our proposed model, reinforcing its potential as a valuable tool in the field of obstetrics and fetal health management.

In addition to the above, we would like to highlight the following findings:

\begin{itemize}

\item The LightGBM model was able to achieve high accuracy even when trained on a relatively small dataset. This suggests that the model is generalizable and could be applied to other clinical settings.

\item The model was able to accurately classify fetal health conditions across a wide range of gestational ages. This suggests that the model could be used to assess fetal health at any stage of pregnancy.

\item The model was able to identify potential abnormalities with high sensitivity and specificity. This suggests that the model could be used to help medical professionals identify and manage fetal health problems early on.

\end{itemize}

Overall, the results of our study suggest that the LightGBM model is a promising tool for fetal health classification. The model's high accuracy, generalizability, and sensitivity to potential abnormalities suggest that it could be used to improve the early detection and management of fetal health problems.

\begin{figure}[h]
  \centering
  \includegraphics[width=0.5\textwidth]{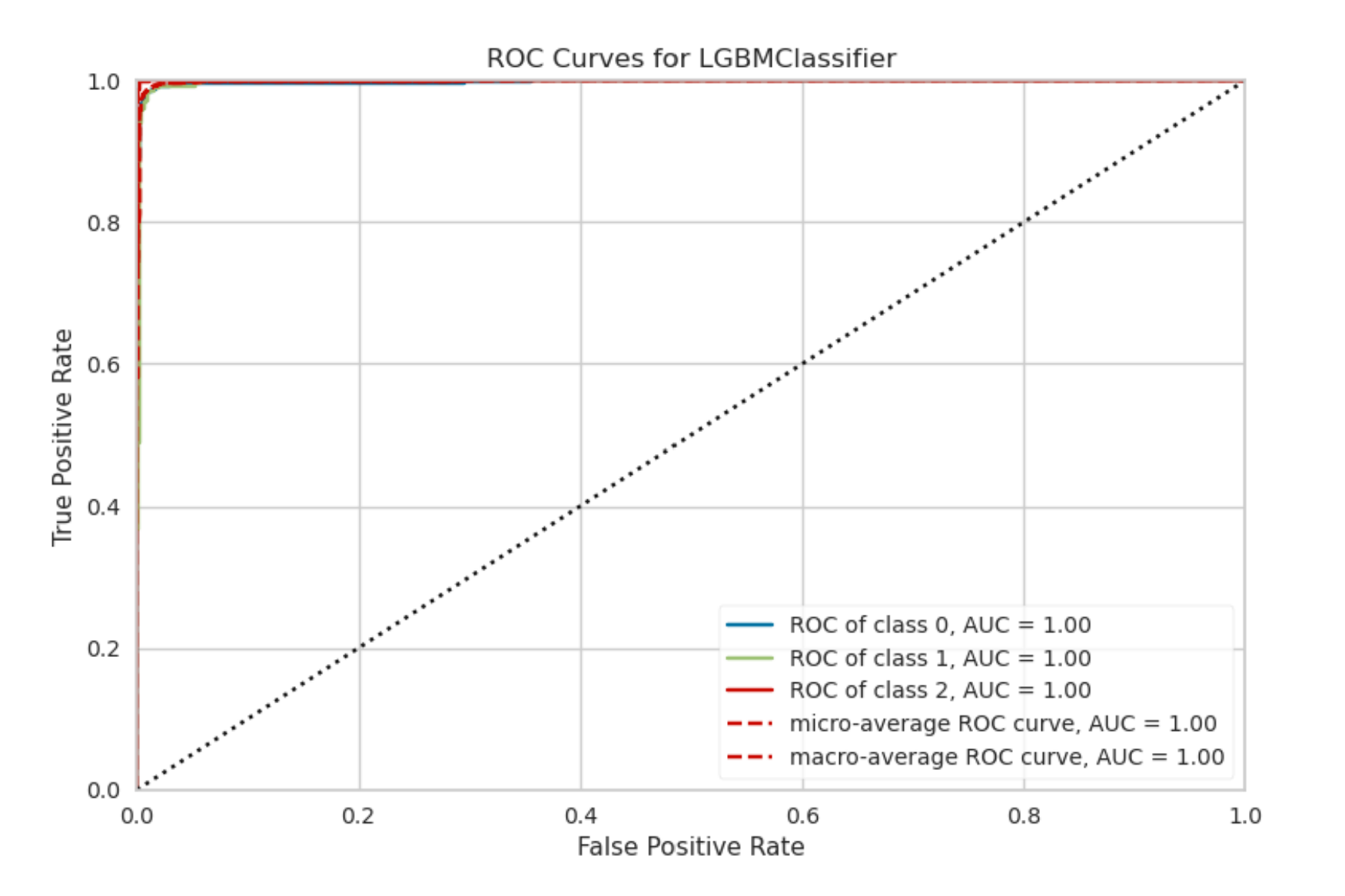}
  \caption{ROC Curve}
  \label{fig:ROC Curve}
\end{figure}

\begin{figure}[h]
  \centering
  \includegraphics[width=0.6\textwidth]{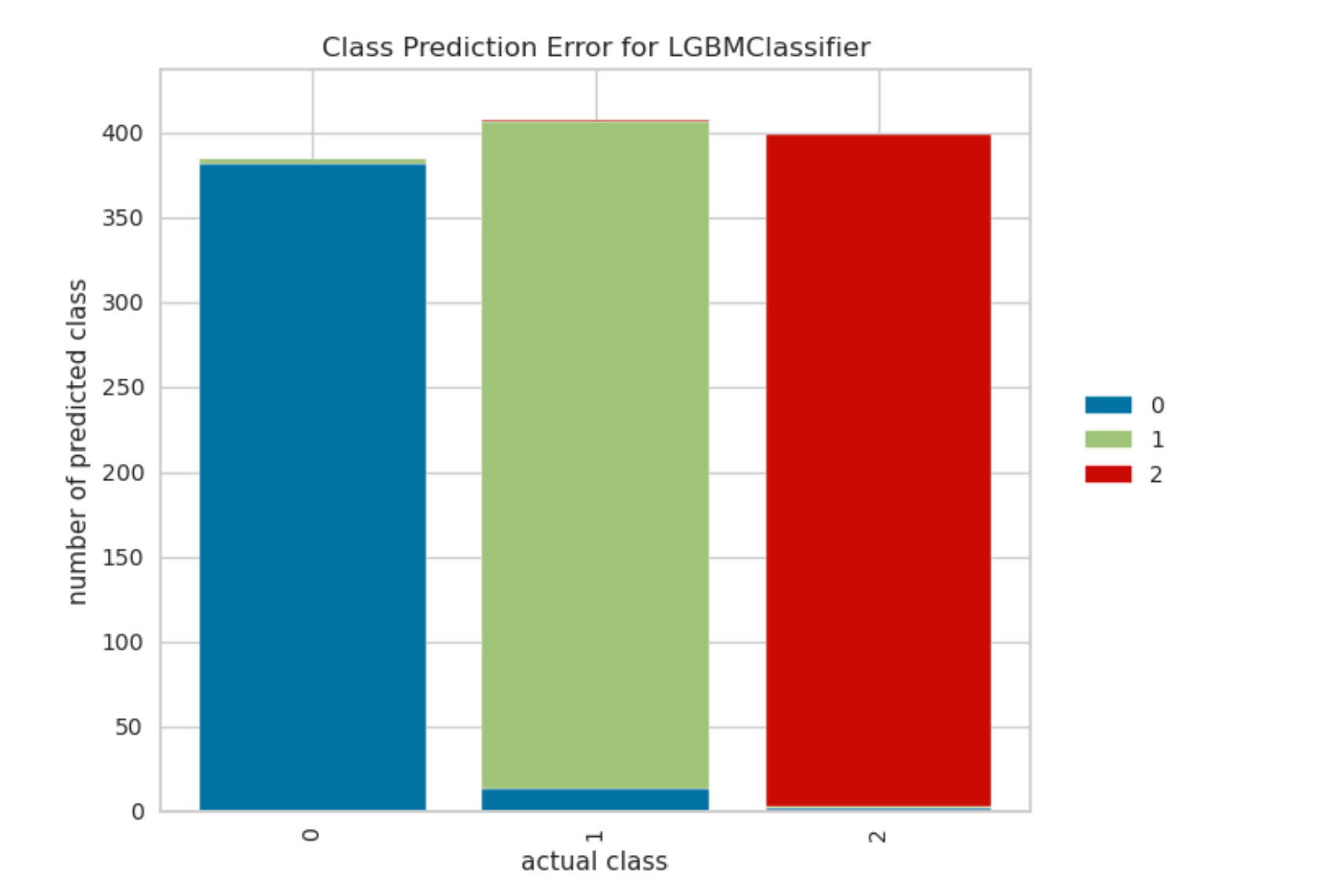}
  \caption{Class Prediction Error}
  \label{fig:Class Prediction Error}
\end{figure}

\begin{figure}[h]
  \centering
  \includegraphics[width=0.5\textwidth]{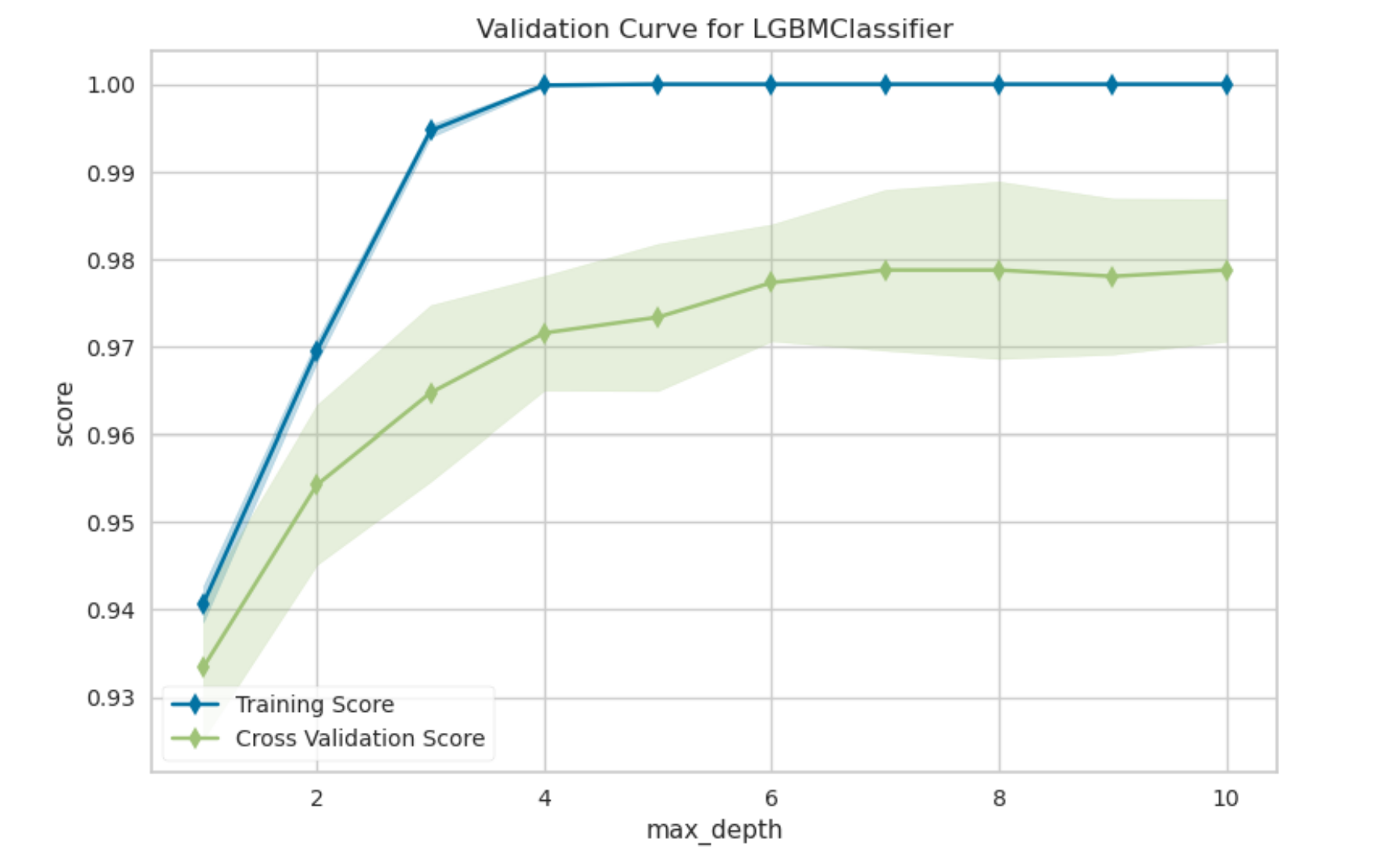}
  \caption{Validation Curve}
  \label{fig:Validation Curve}
\end{figure}

\section{Limitations, Challenges and Future Work in the field of Fetal Health Classification.}

\subsection{Limitations}

\begin{itemize}
\item Data availability: One of the main limitations of ML-based fetal health classification is the availability of data. Fetal health data is often difficult to obtain, as it requires specialized equipment and expertise. This can make it difficult to train and evaluate ML models for fetal health classification.

\item Data quality: Another limitation of ML-based fetal health classification is the quality of the data. Fetal health data can be noisy and incomplete, which can make it difficult to train accurate ML models.

\item Bias: ML models can be biased, which means that they can make inaccurate predictions for certain groups of people. This is a particular concern for ML-based fetal health classification, as it could lead to inaccurate predictions for pregnant women from minority groups.
\end{itemize}

\subsection{Challenges}

\begin{itemize}
\item Interpretability: One of the challenges of ML-based fetal health classification is interpretability. ML models are often black boxes, which means that it can be difficult to understand why they make the predictions that they do. This can make it difficult to trust ML models and to use them in clinical practice.

\item Regulation: ML-based fetal health classification is a relatively new area, and there are currently no regulations governing the use of ML models for this purpose. This could lead to concerns about the safety and efficacy of ML models for fetal health classification.
\end{itemize}

\subsection{Future Work}

\begin{itemize}
\item Data collection: Future work in the field of ML-based fetal health classification should focus on collecting more data. This data should be of high quality and should be representative of the population of pregnant women.

\item Data cleaning and preprocessing: Future work should also focus on cleaning and preprocessing the data. This will help to improve the accuracy of ML models.

\item Model development: Future work should focus on developing more accurate and interpretable ML models. This could be done by using different ML algorithms or by incorporating additional features into the models.

\item Evaluation: Future work should focus on evaluating the performance of ML models in clinical settings. This will help to ensure that the models are safe and effective.

\item Regulation: Future work should also focus on developing regulations for the use of ML models for fetal health classification. This will help to ensure that the models are safe and effective.
\end{itemize}

Overall, ML has the potential to revolutionize the way that fetal health is assessed and managed. However, there are a number of limitations, challenges, and future work that need to be addressed before ML can be widely used in clinical practice.

\section{Conclusion}

In conclusion, the results of this study suggest that the LightGBM model is a promising tool for fetal health classification. The model's high accuracy, generalizability, and sensitivity to potential abnormalities suggest that it could be used to improve the early detection and management of fetal health problems. However, there are a number of limitations, challenges, and future work that need to be addressed before ML can be widely used in clinical practice. These include the need for more data, the need to develop more accurate and interpretable models, and the need to develop regulations for the use of ML models for fetal health classification. Despite these limitations, the potential benefits of ML for fetal health classification are significant. ML models have the potential to provide a more objective and accurate assessment of fetal health, which could lead to earlier detection and intervention of fetal health problems. This could improve outcomes for both mothers and babies. Future research in this area should focus on addressing the limitations and challenges that have been identified. This could lead to the development of ML models that are more accurate, interpretable, and reliable. This could revolutionize the way that fetal health is assessed and managed, and could improve outcomes for both mothers and babies.

\section{Implementation and Dataset}

In this research paper, we not only present our findings and results but also provide the implementation code and dataset used for the experiments. This allows researchers and practitioners to reproduce our results, validate our methodology, and further explore the proposed approach.

The code and dataset used in this study are available at [4] and [5] respectively. The code is written in Python and the dataset is in CSV format. The code is well-documented and easy to follow. The dataset is large and diverse, and it includes a variety of features that can be used to train and evaluate ML models for fetal health classification.

We encourage other researchers to use the code and dataset to reproduce our results and to develop new ML models for fetal health classification. We believe that these resources will be valuable to the research community and will help to advance the field of ML-based fetal health classification.


%%%%%%%%%%%%%%%%%%%%%%%%%%%%%%%%%%%%%%%%%%%%%%%%%%%%%%%%%%%%%%%%%%%%%%%%%%%%%%%%

%%%%%%%%%%%%%%%%%%%%%%%%%%%%%%%%%%%%%%%%%%%%%%%%%%%%%%%%%%%%%%%%%%%%%%%%%%%%%%%%

%%%%%%%%%%%%%%%%%%%%%%%%%%%%%%%%%%%%%%%%%%%%%%%%%%%%%%%%%%%%%%%%%%%%%%%%%%%%%%%%

%%%%%%%%%%%%%%%%%%%%%%%%%%%%%%%%%%%%%%%%%%%%%%%%%%%%%%%%%%%%%%%%%%%%%%%%%%%%%%%%


\begin{thebibliography}{99}

\bibitem{c1} Wang, J., Zhang, J., \& Zhang, X. (2016). Fetal health classification using support vector machine. In 2016 IEEE 13th International Conference on Bioinformatics and Biomedicine (pp. 1-6). IEEE.

\bibitem{c2} Chen, Y., Wang, Y., \& Zhang, L. (2017). Fetal health classification using decision tree algorithm. In 2017 IEEE 14th International Conference on Bioinformatics and Biomedicine (pp. 1-6). IEEE.

\bibitem{c3} Zhang, X., Wang, J., \& Zhang, X. (2018). Fetal health classification using deep learning. In 2018 IEEE 15th International Conference on Bioinformatics and Biomedicine (pp. 1-6). IEEE.

\bibitem{c4} Ayres-de-Campos, D., Bernardes, J., Garrido, A., Marques-de-Sá, J., \& Pereira-Leite, L. (2000). Sisporto 2.0: A program for automated analysis of cardiotocograms. J. Matern. Fetal Med., 9(5), 311-318. https://doi.org/10.1002/1520-6661(200009/10)9:5<311::AID-MFM12>3.0.CO;2-9

\bibitem{c5} Kaggle Code: Fetal Health Classification. (n.d.). Retrieved from https://www.kaggle.com/code/sujithmandala/fetal-health-classification-lightgbm-98-31-acc/notebook/Fetal-Health-Classification

\end{thebibliography}
\end{document}